\pdfoutput=1

\documentclass[11pt]{article}

\usepackage{ACL}
\usepackage{booktabs}

\usepackage{times}
\usepackage{latexsym}
\usepackage{graphicx}
\usepackage{multirow}
\usepackage{pifont}
\usepackage{float}
\newcommand{\cmark}{\ding{51}}%
\newcommand{\xmark}{\ding{55}}%

\usepackage[T1]{fontenc}

\usepackage[utf8]{inputenc}

\usepackage{microtype}

\usepackage{inconsolata}

\usepackage{enumitem}
\usepackage{amsmath}
\DeclareMathOperator*{\argmax}{argmax}

\usepackage{comment}

\usepackage[capitalise,noabbrev]{cleveref}

\title{Parameter-Efficient Fine-Tuning of LLaMA for the Clinical Domain}

\author{
Aryo Pradipta Gema\textsuperscript{1} \quad Pasquale Minervini\textsuperscript{1} \quad Luke Daines\textsuperscript{2} \\ \textbf{Tom Hope\textsuperscript{3,4}} \quad \textbf{Beatrice Alex\textsuperscript{5,6}} \\
\textsuperscript{1}School of Informatics, University of Edinburgh \quad 
\textsuperscript{2}Usher Institute, University of Edinburgh\\
\textsuperscript{3}Allen Institute of AI\\
\textsuperscript{4}Hebrew University of Jerusalem\\
\textsuperscript{5}Edinburgh Futures Institute, University of Edinburgh\\
\textsuperscript{6}School of Literatures, Languages and Cultures, University of Edinburgh\\
\texttt{ \{aryo.gema, p.minervini, luke.daines, b.alex\}@ed.ac.uk}\\
\texttt{tomh@allenai.org}
}

\begin{document}
\maketitle
\begin{abstract}
Adapting pretrained language models to novel domains, such as clinical applications, traditionally involves retraining their entire set of parameters. Parameter-Efficient Fine-Tuning (PEFT) techniques for fine-tuning language models significantly reduce computational requirements by selectively fine-tuning small subsets of parameters. 
In this study, we propose a two-step PEFT framework and evaluate it in the clinical domain.
Our approach combines a specialised PEFT adapter layer designed for clinical domain adaptation with another adapter specialised for downstream tasks. We evaluate the framework on multiple clinical outcome prediction datasets, comparing it to clinically trained language models.
Our framework achieves a better AUROC score averaged across all clinical downstream tasks compared to clinical language models. In particular, we observe large improvements of 4-5\% AUROC in large-scale multilabel classification tasks, such as diagnoses and procedures classification.
To our knowledge, this study is the first to provide an extensive empirical analysis of the interplay between PEFT techniques and domain adaptation in an important real-world domain of clinical applications.\footnote{The code is accessible via \url{https://github.com/aryopg/clinical_peft}.}


\end{abstract}

\section{Introduction}

\begin{figure}[h]
    \centering
    \includegraphics[width=\linewidth,clip]{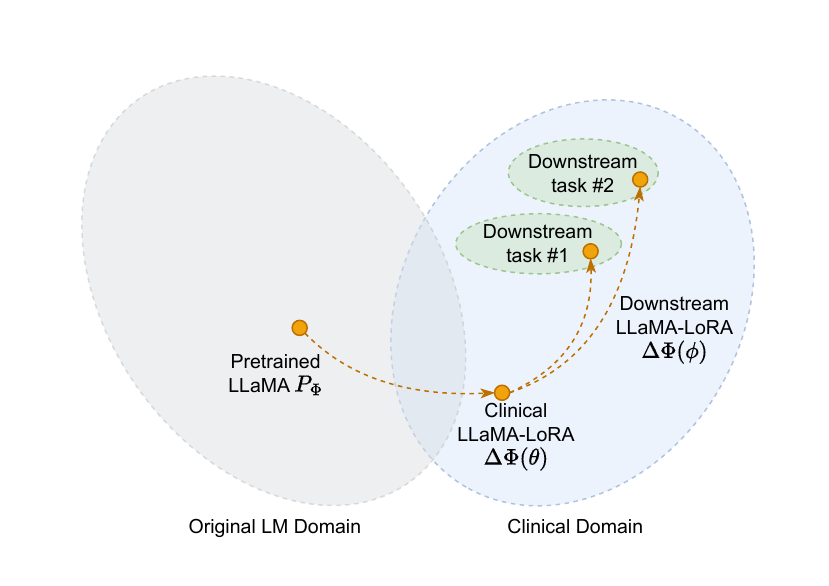}
    \caption{
    An illustration of the proposed two-step PEFT framework. Clinical LLaMA-LoRA fine-tunes the pretrained LLaMA to the clinical domain. Downstream LLaMA-LoRA further fine-tunes the domain-adapted model to downstream clinical tasks.
    }
    \label{fig:clinical_llama_lora}
\end{figure}

Large Language Models (LLMs) have consistently achieved state-of-the-art performance across various NLP tasks.
However, while these models exhibit impressive generalisation abilities, they often struggle to perform in specialised domains such as clinical applications, primarily due to the absence of domain-specific knowledge.
The complexity of medical terminology and the presence of incomplete sentences in clinical notes contribute to this challenge~\cite{lehmanericClinicalT5LargeLanguage}.
Unfortunately, studies have indicated that even LLMs pretrained with datasets comprising biomedical publications still exhibit suboptimal performance when applied to downstream clinical applications, particularly when compared to LLMs pretrained with clinical notes~\cite{alsentzerPubliclyAvailableClinical2019, liClinicalLongformerClinicalBigBirdTransformers2022, yangGatorTronLargeClinical2022}.
This observation suggests that there are intrinsic nuances specific to the clinical context that can only be effectively captured if LLMs undergo pretraining using clinical datasets.

The current approach of adapting pretrained LLMs to the clinical domain typically involves fine-tuning the entire model parameters~\cite{alsentzerPubliclyAvailableClinical2019, pengTransferLearningBiomedical2019, vanakenClinicalOutcomePrediction2021a, michalopoulosUmlsBERTClinicalDomain2021, lehmanericClinicalT5LargeLanguage}.
However, due to the rapid increase in the size of LLMs, such a practice demands extensive computational resources, which may not be readily accessible to all researchers.
Consequently, this challenge will further exacerbate the disparity between the resource-rich and resource-constrained research institutions~\cite{ruderModularParameterEfficientFineTuning2022}.

To address the substantial computational demands, studies have proposed various Parameter-Efficient Fine-Tuning (PEFT) techniques.
These techniques present a practical solution by fine-tuning a small subset of additional parameters while keeping the remaining pretrained parameters fixed.
As a result, this strategy significantly alleviates the computational burden while achieving comparable performance to that of full fine-tuning.

In this study, we propose a two-step PEFT framework (see \cref{fig:clinical_llama_lora}).
Firstly, we introduce Clinical LLaMA-LoRA, a Low-Rank Adaptation~\cite[LoRA,][]{huLoRALowRankAdaptation2021} PEFT adapter built upon the open-source Large Language Model Meta AI (LLaMA) \cite{touvronLLaMAOpenEfficient2023}.
Then, we introduce Downstream LLaMA-LoRA, which is trained on top of the pretrained Clinical LLaMA-LoRA.
Downstream LLaMA-LoRA is specifically designed for clinical downstream tasks.
The fusion of the two adapters achieves better performance in clinical NLP downstream tasks compared to clinically trained LLMs while considerably reducing the computational requirements.
%
This study presents the following contributions:
\begin{itemize}[leftmargin=*]
    \item We introduce Clinical LLaMA-LoRA, a PEFT-adapted version of the LLaMA model tailored specifically for the clinical domain.
    \item We provide comparisons of multiple PEFT techniques in terms of language modelling performance based on perplexity score, shedding light on the optimal PEFT techniques for the clinical domain-adaptive pretraining.
    \item We introduce Downstream LLaMA-LoRA, built on top of Clinical LLaMA-LoRA and tailored specifically for the clinical downstream tasks.
    \item We evaluate the proposed mixture of Clinical LLaMA-LoRA and Downstream LLaMA-LoRA on downstream clinical datasets and tasks. Our proposed framework showcases improvements in AUROC scores over the existing clinical LLMs.
\end{itemize}

\begin{figure*}[h]
    \centering
    \includegraphics[width=\linewidth]{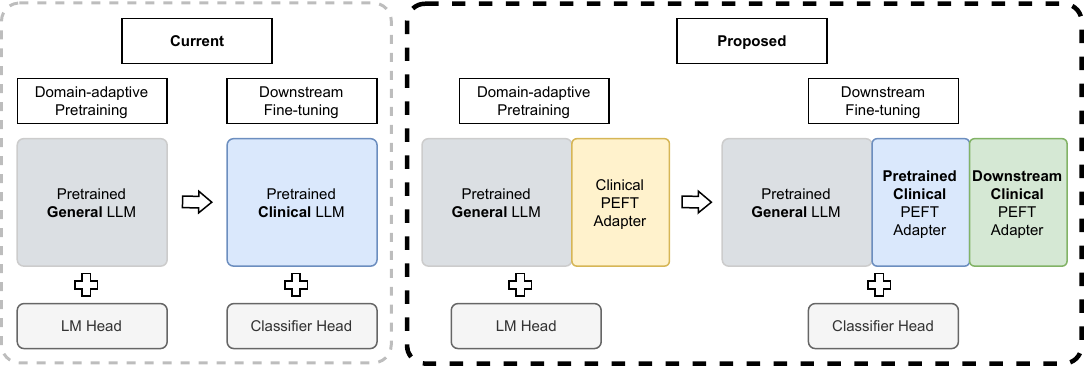}
    \caption{Frameworks of domain-adaptive and downstream fine-tuning to adapt a pretrained LLM from the general domain to the clinical domain. As opposed to a full fine-tuning process which can be prohibitively expensive (left), our approach leverages PEFT techniques to introduce a clinically-specialised adapter that is attached to a pretrained general LLM (right). Our proposed framework also introduces another clinical PEFT adapter trained on the downstream clinical tasks, such as clinical note classification.}
    \label{fig:peft}
\end{figure*}

\section{Background}

\subsection{Biomedical Large Language Models}

General-domain LLMs continue to face challenges when confronted with domain-specific tasks.
The complexity associated with the requisite domain knowledge is recognised as a significant factor~\cite{lingOneModelFitsAllSurveyDomain2023}, particularly within the biomedical domain.
Consequently, numerous studies have attempted to adapt LLMs specifically for the biomedical domain.

An early example of such adaptation is BioBERT~\cite{leeBioBERTPretrainedBiomedical2019}, which was pretrained using biomedical research articles from PubMed and PubMed Central.
This adaptation has shown improved performance across various biomedical NLP tasks.
Recognising the significance of biomedical-specific vocabularies, \citet{guDomainSpecificLanguageModel2022} proposed PubMedBERT, which is pretrained on biomedical data from scratch and initialised the model vocabulary with the biomedical corpus.
The growing interest in biomedical NLP research has led to the adaptation of even larger models to the biomedical domain~\cite{luoBioGPTGenerativePretrained2022, singhalLargeLanguageModels2022, wuPMCLLaMAFurtherFinetuning2023, singhalExpertLevelMedicalQuestion2023}

While these biomedical LLMs have demonstrated advancements in various biomedical NLP benchmarking tasks, studies have revealed that clinical LLMs still outperform their biomedical counterparts in numerous clinical downstream tasks~\cite{alsentzerPubliclyAvailableClinical2019, yangGatorTronLargeClinical2022, liClinicalLongformerClinicalBigBirdTransformers2022, lehmanericClinicalT5LargeLanguage}.
This suggests that domain-adaptive pretraining using clinical data is still the \textit{de facto} protocol in adapting LLMs to the clinical domain.

\subsection{Clinical Large Language Models}

Clinical LLMs are often fine-tuned with clinical data from an LLM that is already pretrained with datasets that encompass broader topics.
For instance, Bio+ClinicalBERT~\cite{alsentzerPubliclyAvailableClinical2019} is domain-adaptively pretrained using clinical notes from the Medical Information Mart for Intensive Care (MIMIC)-III database~\cite{johnsonMIMICIIIFreelyAccessible2016}, starting from a pretrained BioBERT~\cite{leeBioBERTPretrainedBiomedical2019}, which itself is pretrained on biomedical articles.
BlueBERT~\cite{pengTransferLearningBiomedical2019} is domain-adaptively pretrained using PubMed abstracts and MIMIC-III clinical notes from a BERT model~\cite{devlinBERTPretrainingDeep2019}, that is pretrained with general-domain texts.
Similarly, Clinical-T5~\cite{lehmanericClinicalT5LargeLanguage} is domain-adaptively pretrained using the union of MIMIC-III and MIMIC-IV~\cite{johnsonMIMICIVFreelyAccessible2023} clinical notes from T5-base~\cite{raffelExploringLimitsTransfer2020}, another general-domain LLM.

All these studies share a common approach, which is to fine-tune the entire model parameters.
With massive LLMs, this method has become cost-prohibitive and inaccessible for many researchers.

\subsection{Parameter-Efficient Fine-Tuning for Large Language Models}

Suppose that we have a pretrained LLM $P_{\Phi}(y|x)$; fine-tuning it can be effectively defined as finding the most appropriate parameter changes $\Delta \Phi$ by optimising the fine-tuning objective.
A conventional, full fine-tuning process means that the model needs to learn a $\Delta \Phi$ whose dimension is equal to the entire parameters of the pretrained LLM $|\Delta \Phi| = |\Phi_0|$, which is computationally expensive.
PEFT techniques address this by tuning the \emph{delta} $\Delta \Phi$, which corresponds to a very small fraction of additional trainable parameters during the fine-tuning process.

Adapter tuning~\cite{houlsbyParameterEfficientTransferLearning2019} is an early PEFT method that involves adding small additional parameters called \emph{adapters} to each layer of the pretrained model and strictly fine-tuning this small set of new parameters.
LoRA~\cite{huLoRALowRankAdaptation2021} is another PEFT approach that trains low-rank matrices to represent the attention weights update of transformer-based models.

Another group of PEFT approaches leverages the concept of prompting.
Prefix Tuning~\cite{liPrefixTuningOptimizingContinuous2021} optimises a sequence of continuous task-specific vectors, called a \emph{prefix}, which are trainable parameters that do not correspond to real tokens.
P-Tuning~\cite{liuGPTUnderstandsToo2021} uses a similar strategy as Prefix tuning with a focus on text understanding tasks, as opposed to generative tasks.
Prompt tuning~\cite{lesterPowerScaleParameterEfficient2021} simplifies Prefix tuning by introducing trainable tokens, called \emph{soft prompts}, for each downstream task.
\citet{liuPTuningV2Prompt2022} introduced P-tuning v2 which uses deep prompt tuning to address the lack of performance gain in the previous prompt tuning techniques.

By fine-tuning a small fraction of additional parameters, all PEFT approaches alleviate the issue of extensive computational resource requirements.

\subsection{Multi-step Adaptation}

Prior studies have explored the two-step adaptation framework, although they have fundamental differences from our proposed setup.
For instance, \citet{DBLP:journals/corr/abs-2111-00667} introduced a multi-domain unsupervised domain adaptation (UDA) with a two-step strategy, involving domain-fusion training with Masked Language Model loss on a mixed corpus, followed by task fine-tuning with a task-specific loss on the domain corpus.
More recently, \citet{malik-etal-2023-udapter} introduced UDApter which utilises PEFT adapters to do efficient UDA.
However, unsupervised domain matching techniques such as UDApter rely on restrictive assumptions about the underlying data distributions that are often unsatisfied in real-world scenarios~\cite{DBLP:journals/corr/abs-2006-13352}.
In our study, we experiment with the clinical domain as the target domain that is not available in the LLM's initial pretraining.
Consequently, significant discrepancies exist between the distributions of the source and target domains.
Leveraging the amount of available clinical notes, we adopt a self-supervised learning paradigm by continually pretraining the LLMs within the target domain rather than relying on the UDA paradigm.

Our approach shares theoretical similarities with the multi-step continual pretraining approach,  proposed by \citet{gururangan-etal-2020-dont}, which proposes domain- and task-adaptive pretraining.
However, the main difference between our proposed approach and \citet{gururangan-etal-2020-dont} is in the discrepancy between the source and the target domains.
\citet{gururangan-etal-2020-dont} experimented with adapting general-domain LLMs to domains encountered during their initial pretraining, such as news and biomedical domains.
On the other hand, we experiment with the clinical domain which is entirely absent from the LLMs' initial pretraining due to legal constraints which restrict access to sensitive clinical notes.
On top of that, adapting to the clinical domain poses a bigger challenge due to the complexity of medical terminology and the presence of incomplete sentences~\cite{lehmanWeStillNeed2023}.

\section{Methodology}

\subsection{Problem Statement}

\cref{fig:peft} shows the comparison between the current and proposed problem definitions.
The general problem can be decomposed into two stages:

\paragraph{Domain-adaptive Pretraining.} 
Given a pretrained general LLM $P_{\Phi}(y|x)$ with its parameters $\Phi$ and a training dataset $\mathcal{Z} = \{ (x_i, y_i) \}_{i=1,...,N}$.
To adapt to the new domain, the model needs to update its weight iteratively from its pretrained state $\Phi_0$ to $\Phi = \Phi_0 + \Delta \Phi$.
This process of maximising the objective function can be defined as:
\begin{equation*}
    \argmax _{\Phi} \sum_{(x, y) \in \mathcal{Z}} \sum_{t=1}^{|y|} \log \left(P_{\Phi}\left(y_t \mid x, y_{<t}\right)\right)
\end{equation*}
In the current paradigm, a full fine-tuning process means that the model needs to learn a $\Delta \Phi$ whose dimension is equal to the entire pretrained parameters $|\Delta \Phi| = |\Phi_0|$, which is computationally expensive.

In the proposed paradigm, we tune only small additional parameters $\theta$ such that $\Phi = \Phi_0 + \Delta \Phi(\theta)$ whose dimension is very small compared to the original parameters $|\theta| \ll |\Phi_0|$.
Thus, the 
training objective can be redefined as:
\begin{equation*}  
    \argmax _{\theta} \sum_{(x, y) \in \mathcal{Z}} \sum_{t=1}^{|y|} \log \left(P_{\Phi_0 + \Delta \Phi(\theta)}\left(y_t \mid x, y_{<t}\right)\right)
\end{equation*}
In the current paradigm, the outcome of domain-adaptive pretraining would be a clinically-adapted LLM.
While in the proposed paradigm, the outcome would be the clinical PEFT component, which can be combined with the untouched pretrained general LLM for downstream applications.

\paragraph{Downstream Fine-tuning.}
In the current paradigm, the pretrained clinical LLM is fine-tuned to the downstream tasks, such as document classification tasks. Suppose that we have a pretrained clinical LLM $P_{\Phi, \Theta}$ with its domain-adapted parameters $\Phi$ and a newly initialised classifier layer $\Theta$, as well as a training dataset $\mathcal{Z} = \{ (x_i, y_i) \}_{i=1,...,N}$. We want to maximise a specific loss function, such as a cross-entropy loss:
\begin{equation*}
    \argmax _{\Phi, \Theta} \frac{1}{N} \sum_{i=1}^N y_{i} \log \left(P_{\Phi, \Theta}\left(x_{i}\right)\right)
\end{equation*}
In contrast, in the proposed paradigm, the fine-tuning process only updates the small additional parameters $\Delta \Phi(\theta)$ and the classifier head $\Theta$:
\begin{equation*}
    \argmax _{\theta, \Theta} \frac{1}{N} \sum_{i=1}^N y_{i} \log \left(P_{\Phi + \Delta \Phi(\theta), \Theta}\left(x_{i}\right)\right)
\end{equation*}
In fact, we can also decompose the fine-tuning into an additional "delta-updating" process:
\begin{equation*}
    \argmax _{\theta, \phi, \Theta} \frac{1}{N} \sum_{i=1}^N y_{i} \log \left(P_{\Phi + \Delta \Phi(\theta)+ \Delta \Phi(\phi), \Theta}\left(x_{i}\right)\right)
\end{equation*}
Similar to the Domain-adaptive Pretraining stage, the dimensions of the additional parameters $\theta$ and $\phi$ are very small compared to the original parameters.
By updating only the additional parameters and the classifier head, the proposed paradigm reduces the computational requirements, making it more efficient and feasible, especially for clinical settings that are often resource-constrained.

\subsection{Two-step LLaMA-LoRA}

In this study, we propose a two-step PEFT framework (as shown on the right-hand side of~\cref{fig:peft}).
Firstly, we introduce Clinical LLaMA-LoRA, a LoRA adapter built upon LLaMA~\cite{touvronLLaMAOpenEfficient2023} that is adapted to the clinical domain.
Secondly, we introduce Downstream LLaMA-LoRA, which is trained on top of the pretrained Clinical LLaMA-LoRA and is specifically adapted to the downstream tasks.

\paragraph{LLaMA models}

In this study, we evaluate two LLaMA models; the 7 billion parameters version of LLaMA~\cite{touvronLLaMAOpenEfficient2023} and the 7 billion parameters version of PMC-LLaMA\cite{wuPMCLLaMAFurtherFinetuning2023}.
LLaMA was pretrained with an array of texts from multiple sources, such as English CommonCrawl, Wikipedia, ArXiv, and C4~\cite{raffelExploringLimitsTransfer2020}.
While, PMC-LLaMA is a domain-adapted LLaMA model that was pretrained on 4.8 million biomedical academic papers from PubMed Central.

\paragraph{Domain-adaptive Pretraining: Clinical LLaMA-LoRA}

Clinical LLaMA-LoRA is trained using a combination of MIMIC-IV de-identified discharge summaries (331,794) and radiology reports (2,321,355), resulting in a collection of 2,653,149 individual clinical notes.
We evaluate five PEFT techniques, which include \emph{LoRA}~\cite{huLoRALowRankAdaptation2021}, \emph{Adaptation Prompt}~\cite{zhangLLaMAAdapterEfficientFinetuning2023}, \emph{Prefix Tuning}~\cite{liPrefixTuningOptimizingContinuous2021}, \emph{Prompt Tuning}~\cite{lesterPowerScaleParameterEfficient2021}, and \emph{P-tuning}~\cite{liuGPTUnderstandsToo2021}.

Our approach follows the autoregressive language modelling pretraining objective employed in the original LLaMA training.
To ensure compatibility with available computational resources, we use fixed model hyperparameters that allow us to fit the LLM into a single NVIDIA A100-80GB GPU (see Appendix~\ref{app:llama_pretraining_hyperparams}).
We optimise the hyperparameters specific to each PEFT method using Gaussian Process regression for Bayesian Optimisation~\citep{DBLP:journals/corr/abs-1807-02811}~\footnote{Specifically, we use the W\&B Sweep APIs: \url{https://docs.wandb.ai/guides/sweeps}}
with a maximum of 20 trials.
The detailed hyperparameters search space can be found in Appendix~\ref{app:llama_pretraining_hpo_space}.
During this stage, we evaluate the perplexity scores of the LLM variants.

\paragraph{Downstream Fine-tuning: Downstream LLaMA-LoRA}

\begin{table}[t]
    \centering
    \small
    \setlength\tabcolsep{3.75pt}
    \begin{tabular}{lccccc}
        \toprule
        {\bf Dataset} & {\bf \# Class} & {\bf Multilabel} & {\bf \# Train} & {\bf \# Valid} & {\bf \# Test} \\
        \midrule
        LOS & 4 & \xmark & 30,421 & 4,391 & 8,797 \\
        MOR & 2 & \xmark & 33,954 & 4,908 & 9,822 \\
        PMV & 2 & \xmark & 5,666 & 707 & 706 \\
        DIAG & 1,266 & \cmark & 33,994 & 4,918 & 9,829 \\
        PROC & 711 & \cmark & 30,030 & 4,357 & 8,681 \\
        \bottomrule
    \end{tabular}
    \caption{Statistics and types of downstream clinical document classification tasks: length of stay (LOS), mortality (MOR), prolonged mechanical ventilation (PMV), diagnoses (DIAG), and procedures (PROC).}
    \label{tab:downstream_tasks_stats}
\end{table}

We fine-tune the Clinical LLaMA-LoRA and Downstream LLaMA-LoRA to clinical document classification tasks:
\begin{itemize}[itemsep=0pt, parsep=0pt, leftmargin=*]
    \item \textbf{Prolonged mechanical ventilation (PMV)}: a binary classification task to predict whether a patient will require mechanical ventilation for more than seven days~\cite{huangClinicalXLNetModeling2020, naikLiteratureAugmentedClinicalOutcome2022d}.
    \item \textbf{In-hospital mortality (MOR)}: a binary classification task to predict whether a patient will survive during their hospital stay~\cite{vanakenClinicalOutcomePrediction2021a, naikLiteratureAugmentedClinicalOutcome2022d}.
    \item \textbf{Length of stay (LOS)}:  a multiclass classification task to predict the length of a patient's hospital stay, categorised into four time-bins: less than three days, three to seven days, one to two weeks, and more than two weeks~\cite{vanakenClinicalOutcomePrediction2021a, naikLiteratureAugmentedClinicalOutcome2022d}.
    \item \textbf{Diagnoses (DIAG)}: a large-scale multilabel classification task to predict the differential diagnoses of a patient, represented by simplified ICD-9 diagnosis codes~\cite{vanakenClinicalOutcomePrediction2021a}.
    \item \textbf{Procedures (PROC)}: a large-scale multilabel classification task to predict the treatments administered to a patient, represented by simplified ICD-9 procedure codes~\cite{vanakenClinicalOutcomePrediction2021a}.
\end{itemize}
The label and split statistics of each dataset can be found in~\cref{tab:downstream_tasks_stats}.

During this downstream fine-tuning process, we use fixed model hyperparameters to ensure compatibility with the available computational resources, a single NVIDIA A100-80GB GPU (see Appendix~\ref{app:downstream_hyperparams}).
We optimise the hyperparameters specific to each PEFT method using Gaussian Process regression for Bayesian Optimisation with a maximum of 20 trials.
The detailed hyperparameters search space of the PEFT method can be found in Appendix~\ref{app:downstream_hpo_space}.

For evaluating the performance of the model on these downstream tasks, we report the Area Under the Receiver Operating Characteristic Curve (AUROC) scores. Additionally, we report the macro-averaged AUROC score across all clinical tasks as commonly done in NLP benchmarking tasks~\cite{DBLP:conf/iclr/WangSMHLB19, pengTransferLearningBiomedical2019, guDomainSpecificLanguageModel2022}.

\subsection{Baseline Models}

We selected baseline models that have undergone a domain-adaptive pretraining process on clinical notes (MIMIC-III).
Thus, these baseline models have been designed to perform specifically on clinical data, providing comparison points for evaluating our proposed approach of two-step adaptation in downstream clinical NLP tasks.
The baseline models used in the evaluation are as follows:

\begin{itemize}[itemsep=0pt, parsep=0pt, leftmargin=*]
    \item \textbf{Bio+ClinicalBERT}~\cite{alsentzerPubliclyAvailableClinical2019}: Bio+ClinicalBERT is pretrained on MIMIC-III clinical notes. It is initialised from a biomedical language model called BioBERT~\cite{leeBioBERTPretrainedBiomedical2019}, which is pretrained on biomedical research articles.
    \item \textbf{BlueBERT}~\cite{pengTransferLearningBiomedical2019}: BlueBERT is pretrained on MIMIC-III clinical notes and PubMed abstracts starting from the pretrained checkpoint of BERT~\cite{devlinBERTPretrainingDeep2019}, a general-domain language model.
    \item \textbf{CORe}~\cite{vanakenClinicalOutcomePrediction2021a}: CORe is pretrained on MIMIC-III clinical notes and biomedical articles starting from the pretrained checkpoint of BioBERT~\cite{leeBioBERTPretrainedBiomedical2019}.
    \item \textbf{UmlsBERT}~\cite{michalopoulosUmlsBERTClinicalDomain2021}: UmlsBERT is pretrained on MIMIC-III clinical notes using the pretrained weights of Bio+ClinicalBERT with modified architecture and pretraining objective that incorporates knowledge from the Unified Medical Language System (UMLS) Metathesaurus~\cite{schuylerUMLSMetathesaurusRepresenting1993}.
\end{itemize}


\section{Results and Analysis}

\subsection{Domain-adaptive Pretraining}

\begin{table*}[htbp]
    \centering
    \small
    \begin{tabular}{clcccccc}
        \toprule
        \textbf{Base Model} & \textbf{PEFT} & \textbf{Trainable Params} & \textbf{Train Ppl} & \textbf{Test Ppl} & \textbf{GPU} & \textbf{Train Time (h:m:s)} \\
        \midrule
        Clinical LLaMA & - & 6.7B (100\%) & 1.811 & 2.210 & 4x80GB & 49:26:38 \\
        \midrule
        \multirow{5}{*}{LLaMA} & \textbf{LoRA} & \textbf{8.4M (0.12\%)} & \textbf{1.858} & \textbf{2.244} & 1x80GB & \textbf{21:37:42} \\
        & Adaptation Prompt & 1.2M (0.02\%) & 2.561 & 2.865 & 1x80GB & 24:57:17 \\
        & Prefix Tuning & 5.2M (0.08\%) & 2.815 & 2.748 & 1x80GB & 20:11:07 \\
        & Prompt Tuning & 61.4K (0.0009\%) & 4.846 & 4.007 & 1x80GB & 23:27:28 \\
        & P-tuning & 16.1M (0.24\%) & 2.723 & 3.271 & 1x80GB & 23:49:31 \\
        \midrule
        \multirow{5}{*}{PMC-LLaMA} & \textbf{LoRA} & \textbf{2.1M (0.03\%)} & \textbf{1.938} & \textbf{2.404} & 1x80GB & \textbf{21:32:59} \\
        & Adaptation Prompt & 1.2M (0.018\%) & 2.374 & 2.867 & 1x80GB & 23:33:10 \\
        & Prefix Tuning & 2.6M (0.04\%) & 1.789 & 2.848 & 1x80GB & 20:13:10 \\
        & Prompt Tuning & 41K (0.0006\%) & 4.821 & 4.385 & 1x80GB & 22:25:32 \\
        & P-tuning & 2.2M (0.03\%) & 3.491 & 4.572 & 1x80GB & 22:28:15 \\
        \bottomrule
    \end{tabular}
    \caption{
    Domain-adaptive Pretraining results of LLaMA and PMC-LLaMA trained on MIMIC-IV clinical notes with a language modelling objective.
    Lower perplexity scores indicate better language modelling performance.
    The \textbf{boldface row} indicates the model with the lowest perplexity score from each base model variant.
    }
    \label{tab:pretraining_results}
\end{table*}

The pretraining results can be found in~\cref{tab:pretraining_results}.
We employ PEFT techniques for domain-adaptive pretraining, requiring a significantly smaller number of parameters ranging from just 0.001\% to 0.24\% of the original model parameters.
This approach substantially reduces the required computational resources and training time.
We perform a full-parameter domain-adaptive pretraining of LLaMA, referred to as \textbf{Clinical LLaMA}, using four NVIDIA A100-80GB GPUs which took 49.5 hours.
Instead, PEFT techniques require less than 24 hours per epoch on average with only a single GPU with a comparable perplexity score.

LoRA emerges as the best-performing PEFT method for both LLaMA and PMC-LLaMA in the clinical domain-adaptive pretraining, achieving the lowest perplexity scores of 2.244 and 2.404, respectively, which are very similar to Clinical LLaMA's perplexity score of 2.210.
This pretrained LoRA is referred to as \textbf{Clinical LLaMA-LoRA} in the subsequent sections.
The following experiments in downstream fine-tuning will utilise this pretrained Clinical LLaMA-LoRA.

\subsection{Downstream Fine-tuning}

\begin{table*}[h]
    \centering
    \small
    \begin{tabular}{lcccccc}
    \toprule
    \textbf{Model}                                          & \textbf{PMV}    & \textbf{MOR}    & \textbf{LOS}    & \textbf{DIAG}   & \textbf{PROC}   & \textbf{Macro Average}      \\
    \midrule
    BlueBERT                                       & 57.31 & 81.34 & 72.92 & 73.39 & 76.62 & 72.32 \\
    \textit{UmlsBERT}                              & \textit{\textbf{58.29}} & \textit{81.83} & \textit{73.02} & \textit{72.08} & \textit{78.32} & \textit{72.70} \\
    Bio+ClinicalBERT                               & 54.00 & 72.67 & 72.21 & 76.65 & 83.21 & 71.75 \\
    CORe                                           & 52.11 & 71.52 & 64.17 & 72.40 & 84.51 & 69.40 \\
    \midrule
    Clinical LLaMA* & 52.28 & 63.22 & 56.06 & 59.31 & 63.42 & 58.86 \\
    \midrule
    LLaMA$\ast$                                    & 51.38 & 66.80 & 57.65 & 60.06 & 63.83 & 58.61 \\
    \hspace{5mm}+ LoRA                             & 51.65 & 74.89 & 65.70 & 78.37 & 87.49 & 71.62 \\
    \hspace{5mm}+ Clinical LLaMA-LoRA (Frozen)     & 52.22 & 60.88 & 55.05 & 57.64 & 62.48 & 57.65 \\
    \hspace{10mm}+ Downstream LLaMA-LoRA           & 52.31 & 61.72 & 55.16 & 57.70 & 62.58 & 57.90 \\
    \hspace{5mm}+ Clinical LLaMA-LoRA (Trainable)  & 51.41 & 81.16 & 72.44 & \textbf{81.97} & \textbf{88.69} & 75.13 \\
    \hspace{10mm}\textit{+ Downstream LLaMA-LoRA}  & \textit{53.81} & \textit{\textbf{83.02}} & \textit{\textbf{73.26}} & \textit{81.93} & \textit{88.31} & \textit{\textbf{76.07}} \\
    \midrule
    PMC-LLaMA$\ast$                                & 53.06 & 66.77 & 57.94 & 60.17 & 64.63 & 60.51 \\
    \hspace{5mm}\textit{+ LoRA}                    & \textit{53.84} & \textit{78.03} & \textit{66.14} & \textit{78.81} & \textit{86.68} & \textit{72.70} \\
    \hspace{5mm}+ Clinical LLaMA-LoRA (Frozen)     & 51.33 & 67.19 & 58.13 & 63.59 & 68.26 & 60.06 \\
    \hspace{10mm}+ Downstream LLaMA-LoRA           & 50.90 & 67.00 & 58.31 & 60.50 & 64.42 & 60.23 \\
    \hspace{5mm}+ Clinical LLaMA-LoRA (Trainable)  & 52.88 & 75.86 & 65.89 & 79.66 & 86.85 & 72.23 \\
    \hspace{10mm}+ Downstream LLaMA-LoRA           & 52.21 & 76.54 & 68.42 & 78.67 & 87.08 & 72.58 \\
    \bottomrule
    \end{tabular}
    \caption{AUROC scores in clinical downstream document classification tasks. The macro-averaged AUROC score is calculated by taking the average of AUROC scores across all tasks. The \textbf{boldface cell} indicates the highest AUROC score in a column, the \textit{row in italic} indicates the variant with the highest macro-averaged AUROC in its category. \textit{+ LoRA} denotes applying LoRA on top of the pretrained LLM without domain-adaptive pretraining. \textit{+ Clinical LLaMA-LoRA} denotes applying Clinical LLaMA-LoRA that is domain-adaptively pretrained on top of the pretrained LLM. \textit{+ Downstream LLaMA-LoRA}  denotes applying Downstream LLaMA-LoRA on top of the LLM + Clinical LLaMA-LoRA. \textit{Frozen} means that the parameters are not trainable, while \textit{Trainable} means that the parameters are trainable. $\ast$ Due to restricted computing resources, the fine-tunings of Clinical LLaMA, LLaMA, and PMC-LLaMA were constrained to only training the final classification layer.}
    \label{tab:downstream_results}
\end{table*}

From the downstream fine-tuning results shown in~\cref{tab:downstream_results}, we can decompose the analysis into multiple research questions:

\paragraph{Can LoRA help fine-tune LLaMA from other domains (general and biomedical) to achieve higher AUROC scores in clinical tasks?}

We compare the results obtained by LLaMA and LLaMA + LoRA, as well as PMC-LLaMA and PMC-LLaMA + LoRA, as presented in \cref{tab:downstream_results}.
The obtained results consistently demonstrate improved AUROC scores when utilising LoRA across all tasks.
The macro-averaged AUROC score of LoRA-equipped LLaMA shows a notable 13.01\% increase when compared to the LLaMA-only baseline.
Similarly, LoRA-equipped PMC-LLaMA exhibits a 12.19\% improvement in macro-averaged AUROC compared to the original PMC-LLaMA
Both LLaMA and PMC-LLaMA, when equipped with LoRA, show significant AUROC score improvements in all tasks except the PMV prediction task, which is challenging for all model variants.

Furthermore, the marginal difference in AUROC scores between PMC-LLaMA and the general-domain LLaMA may be attributed to two factors.
Firstly, the original LLaMA has been exposed to biomedical concepts during its pretraining, reducing the need for domain-adaptive pretraining to the biomedical domain.
Secondly, clinical outcome prediction requires an understanding of how to apply biomedical knowledge in an interconnected manner to provide prognostic.
We believe that biomedical pretraining may not be sufficient in providing such practical knowledge.

\paragraph{Can LoRA-equipped LLaMA and PMC-LLaMA perform comparably in comparison to clinically trained LMs?}

We compare the AUROC scores obtained by the baseline models, and LoRA-equipped LLaMA and PMC-LLaMA (see \cref{tab:downstream_results}).
Among the baseline models, UmlsBERT performs the best with a macro-averaged AUROC score of 72.70\%.
Compared to UmlsBERT, both LLaMA and PMC-LLaMA underperform with macro-averaged AUROC scores of 58.61\% and 60.51\%, respectively.
This finding highlights the importance of clinical-specific fine-tuning.

Significant improvements can be observed in LoRA-equipped LLaMA and PMC-LLaMA, with macro-averaged AUROC scores of 71.62\% and 72.70\%, respectively, with noticeable improvements in the diagnoses and procedures prediction tasks.
LoRA-equipped LLaMA achieves AUROC scores of 78.37\% and 87.49\% in the diagnoses and procedures prediction tasks, respectively, compared to 72.08\% and 78.32\% for UmlsBERT. This represents improvements of 6.29\% in diagnoses prediction and 9.17\% in procedures prediction.
Improvements are also observed in the results obtained by LoRA-equipped PMC-LLaMA,
outperforming UmlsBERT by 6.73\% in diagnoses prediction and 8.36\% in procedures prediction.

\paragraph{Can LLaMA and PMC-LLaMA with Clinical LLaMA-LoRA achieve higher AUROC scores than the clinically trained LMs?}

The domain-adaptive pretraining step yields the clinically-trained LoRA adapters for LLaMA and PMC-LLaMA, denoted as \textbf{Clinical LLaMA-LoRA}.
We compare the results of Clinical LLaMA-LoRA-equipped LLaMA and PMC-LLaMA with the baseline models.
We evaluate Clinical LLaMA-LoRA with and without fine-tuning, referred to as "Trainable" and "Frozen" respectively.

The results indicate that Clinical LLaMA-LoRA-equipped LLaMA and PMC-LLaMA outperform the baseline models.
LLaMA with a trainable Clinical LLaMA-LoRA achieves an AUROC score of 75.13\%, surpassing UmlsBERT's score of 72.32\%.
PMC-LLaMA with a trainable Clinical LLaMA-LoRA achieves a lower AUROC score of 72.23\%.
LLaMA with a trainable Clinical LLaMA-LoRA also outperforms Clinical LLaMA which achieves an AUROC score of 58.86\%.

These findings indicate that the Clinical LLaMA-LoRA contributes to higher AUROC scores for LLaMA and PMC-LLaMA over clinically trained LLMs, while biomedical domain-adaptive pretraining may not be necessary to improve the model's performance in the clinical settings.

\paragraph{Can LLaMA and PMC-LLaMA with Clinical LLaMA-LoRA achieve higher AUROC scores than the other fine-tuning variants?}

We examine the importance of the domain-adapted LoRA by comparing the results obtained by LLaMA and PMC-LLaMA equipped with Clinical LLaMA-LoRA against the results of LLaMA and PMC-LLaMA fine-tuning, both original and with LoRA.

Firstly, we evaluate the frozen pretrained Clinical LLaMA-LoRA.
Both LLaMA and PMC-LLaMA with frozen Clinical LLaMA-LoRA do not exhibit a significant increase in performance compared to the original fine-tuning.
This indicates that, despite the domain-adaptive pretraining, the limited number of trainable parameters during the downstream fine-tuning restricts the potential improvement that the model can achieve.
A similar finding can also be observed in the Clinical LLaMA fine-tuning whose overall performance does not differ from the original fine-tuning.
This finding is further supported by the improvement in the AUROC scores of LLaMA and PMC-LLaMA with trainable Clinical LLaMA-LoRA, which achieve 75.13\% and 72.23\% macro-averaged AUROC scores, respectively. These represent substantial improvements from the vanilla fine-tuning performance, 58.61\% and 60.51\% AUROC scores.

\paragraph{Can a downstream LoRA adapter improve the AUROC scores of LLaMA and PMC-LLaMA equipped with Clinical LLaMA-LoRA?}

By considering Clinical LLaMA-LoRA as the "delta-updating" outcome of the domain-adaptive pretraining, we can view the downstream fine-tuning process as an additional "delta-updating" step.
To investigate the impact of this approach, we conduct experiments by adding a Downstream LLaMA-LoRA to LLaMA and PMC-LLaMA models that were already equipped with Clinical LLaMA-LoRA.
From \cref{tab:downstream_results}, we can observe that Downstream LLaMA-LoRA fails to improve the performance of LLaMA and PMC-LLaMA with frozen Clinical LLaMA-LoRA.
On the other hand, improvement can be observed when adding Downstream LLaMA-LoRA to LLaMA with trainable Clinical LLaMA-LoRA.
This combination of LLaMA with trainable Clinical LLaMA-LoRA and Downstream LLaMA-LoRA achieves the highest macro-averaged AUROC score of 76.07\%.
The macro-averaged AUROC score of Clinical LLaMA-LoRA was almost similar to that of PMC-LLaMA with LoRA, suggesting similar efficacy between Clinical LLaMA-LoRA and the full fine-tuning process that PMC-LLaMA has undergone.
Moreover, Clinical LLaMA-LoRA offers the advantage of reduced computational resources and training time, which is aligned with the requirements of practical implementation in clinical settings.

Overall, our proposed method manages to achieve better performance in comparison to clinically trained models.
We also provide a comparison with the state-of-the-art method of PMV, mortality, and length of stay predictions, called BEEP~\cite{naikLiteratureAugmentedClinicalOutcome2022d}, which leverages retrieval augmentation method to provide more contextual information to the model during inference.
The comparison is only partial as BEEP models were not evaluated on the diagnosis and procedure prediction tasks.
As shown in Appendix~\ref{app:beep_results}, our best-performing model achieves a 70.03\% averaged AUROC score, which is slightly worse compared to the best-performing BEEP model with 72.26\% averaged AUROC score.
However, it is worth noting that our proposed method and the state-of-the-art method are complementary to each other.
Hence, future work may explore the possibility of combining the two approaches.

\section{Conclusions}

In this study, we propose a two-step PEFT framework.
We introduce Clinical LLaMA-LoRA, a LoRA~\cite{huLoRALowRankAdaptation2021} adapter built upon LLaMA~\cite{touvronLLaMAOpenEfficient2023}.
Then, we introduce Downstream LLaMA-LoRA, a task-specific adapter that is trained on top of the pretrained Clinical LLaMA-LoRA.
The fusion of the two adapters achieves an AUROC score of 76.07\% macro-averaged across all clinical NLP downstream tasks, which represents a 3.37\% improvement over the best-performing clinical LLM.
Our proposed framework achieves improvement in performance while reducing the computational requirements, which is suited for clinical settings that are often constrained by their computational power.

\section*{Limitations}

This study presents a two-step PEFT framework aimed at effectively adapting LLMs to diverse clinical downstream applications.
However, the evaluation of our model was restricted to MIMIC-based datasets, which are constrained to English and obtained exclusively within the Commonwealth of Massachusetts, United States of America.
Consequently, despite the promising efficacy demonstrated by our proposed method, it would have been advantageous to directly assess its performance across diverse hospital systems spanning other geographical locations and languages.
This would enable a more comprehensive understanding of its applicability and generalizability.
However, it is essential to acknowledge that conducting such an analysis would require working within a trusted research environment and obtaining the necessary permissions to access the relevant datasets.

It is crucial to recognise the restrictions imposed on accessing internal clinical datasets, as they limit our ability to evaluate the effectiveness of our approach across different care provider systems.
Therefore, we encourage care providers to conduct internal experiments within their trusted research environment to ensure the efficacy of our proposed method within their specific use cases should they adopt this approach.

Despite the demonstrated performance improvements, the proposed model may still be susceptible to spurious correlations.
Predicting patient outcomes solely based on clinical notes presents significant challenges due to the other factors that may not be captured within those notes.
For instance, the length of a patient's in-hospital stay is not solely correlated with their diagnoses and disease progression. Factors such as the patient's insurance status, which is not typically mentioned in clinical notes, can severely impact the duration of a patient's stay.
Therefore, we encourage end users of such clinical LLMs to consider additional measures to ensure predictions that reflect a holistic view of the patient's situation, instead of relying solely on the predictions of LLMs.

\section*{Ethics Statement}


In this study, we use MIMIC-based datasets obtained after completing the necessary training.
These datasets comply with de-identification standards set by the Health Insurance Portability and Accountability Act (HIPAA) through data cleansing.
Due to privacy concerns, we refrain from including direct excerpts of the data in the paper.
We also refrain from publicly sharing the pretrained checkpoints.

While our model demonstrates effectiveness, it is important to acknowledge the risks associated with relying solely on clinical outcome prediction models.
There are crucial pieces of information that can be found beyond the scope of clinical notes.
Considering the potential impact on patient health outcomes, it is crucial to exercise caution when utilising these clinical LLMs.
Therefore, we propose that the PEFT adapter generated by our framework, in conjunction with the pretrained LLM, should be used as an aid rather than a replacement for trained clinical professionals.

\section*{Acknowledgements}
%
APG was supported by the United Kingdom Research and Innovation (grant EP/S02431X/1), UKRI Centre for Doctoral Training in Biomedical AI at the University of Edinburgh, School of Informatics.
PM was partially funded by the European Union’s Horizon 2020 research and innovation programme under grant agreement no. 875160, ELIAI (The Edinburgh Laboratory for Integrated Artificial Intelligence) EPSRC (grant no. EP/W002876/1), an industry grant from Cisco, and a donation from Accenture LLP; and is grateful to NVIDIA for the GPU donations.
BA was partially funded by Legal and General PLC as part of the Advanced Care Research Centre and by the Artificial Intelligence and Multimorbidity: Clustering in Individuals, Space and Clinical Context (AIM-CISC) grant NIHR202639.
For the purpose of open access, the authors have applied a Creative Commons attribution (CC BY) licence to any author-accepted manuscript version arising.
This work was supported by the Edinburgh International Data Facility (EIDF) and the Data-Driven Innovation Programme at the University of Edinburgh.

We also thank Yahan Li, Ajitha Rajan, and Bruce Guthrie for their feedback on the manuscript.

\bibliography{anthology,references,custom}

\clearpage

\appendix

\section{Hyperparameters for the Domain-adaptive Pretraining}

\subsection{Fixed Model Hyperparameters}
\label{app:llama_pretraining_hyperparams}

\begin{table}[H]
    \centering
    \scriptsize
    \begin{tabular}{ll}
        \toprule
        Hyperparameter & Value \\
        \midrule
        Learning rate & 3e-4 \\
        Warmup steps ratio & 0.06 \\
        Maximum sequence length & 512 \\
        Gradient accumulation step & 4 \\
        Batch size & 10 \\
        \bottomrule
    \end{tabular}
    \caption{Fixed model hyperparameters for language modelling pretraining. These hyperparameters remain unchanged to fit LLaMA into a single GPU.}
    \label{tab:llama_pretraining_model_hyperparams}
\end{table}

\subsection{PEFT Hyperparameters Optimisation Search Space}
\label{app:llama_pretraining_hpo_space}

\begin{table}[H]
    \centering
    \scriptsize
    \begin{tabular}{llc}
        \toprule
        PEFT & Hyperparameter & Search space \\
        \midrule
        \multirow{3}{*}{LoRA} & r & [2, 4, 8, 16] \\
                              & alpha & [4, 8, 16, 32] \\
                              & dropout & [0.0, 0.1, 0.2] \\
        \midrule
        \multirow{2}{*}{Prefix Tuning} & num virtual tokens & [1, 5, 10, 15, 20] \\
                                       & prefix projection & [true, false] \\
        \midrule
        \multirow{3}{*}{Prompt Tuning} & num virtual tokens & [1, 5, 10, 15, 20] \\
                                       & prompt init & [text, random] \\
        \midrule
        \multirow{5}{*}{P-Tuning} & num virtual tokens & [1, 5, 10, 15, 20] \\
                                  & reparameterisation & ["MLP", "LSTM"] \\
                                  & hidden size & [64, 128, 256, 768] \\
                                  & num layers & [1, 2, 4, 8, 12] \\
                                  & dropout & [0.0, 0.1, 0.2] \\
        \midrule
        \multirow{2}{*}{Adaptation Prompt} & adapter length & [5, 10] \\
                                                  & adapter layers & [10, 20, 30] \\
        \bottomrule
    \end{tabular}
    \caption{The search space for PEFT Hyperparameters optimisation runs during the domain adaptation fine-tuning with language modelling objective. Each PEFT technique has a specific set of hyperparameters to tune, we selected the combination of hyperparameters which has the lowest perplexity score.}
    \label{tab:llama_pretraining_hpo_space}
\end{table}

Specifically for Prompt Tuning, we use a common prompt initialisation text "Finish this clinical note:".

\section{Hyperparameters for the Downstream Fine-tuning}

\subsection{Fixed Model Hyperparameters}
\label{app:downstream_hyperparams}

\begin{table}[H]
    \centering
    \scriptsize
    \begin{tabular}{ll}
        \toprule
        Hyperparameter & Value \\
        \midrule
        Learning rate & 5e-5 \\
        Warmup steps ratio & 0.06 \\
        Maximum sequence length & 512 \\
        Gradient accumulation step & 10 \\
        Batch size & 10 \\
        \bottomrule
    \end{tabular}
    \caption{Fixed model hyperparameters for the clinical downstream fine-tuning. These hyperparameters remain unchanged to fit LLaMA into a single GPU.}
    \label{tab:downstream_hyperparams}
\end{table}

\subsection{PEFT Hyperparameters Optimisation Search Space}
\label{app:downstream_hpo_space}

\begin{table}[H]
    \centering
    \scriptsize
    \begin{tabular}{llc}
        \toprule
        PEFT & Hyperparameter & Search space \\
        \midrule
        \multirow{3}{*}{LoRA} & r & [2, 4, 8, 16] \\
                              & alpha & [4, 8, 16, 32] \\
                              & dropout & [0.0, 0.1, 0.2] \\
        \bottomrule
    \end{tabular}
    \caption{The search space for PEFT Hyperparameters optimisation runs during the downstream fine-tuning. Each PEFT technique has a specific set of hyperparameters to tune, we selected the combination of hyperparameters which has the highest AUROC score.}
    \label{tab:downstream_hpo_space}
\end{table}

\section{Comparison with BEEP~\cite{naikLiteratureAugmentedClinicalOutcome2022d}}
\label{app:beep_results}

\begin{table}[H]
    \centering
    \small
    \begin{tabular}{lcccc}
    \toprule
    \textbf{Model}                                          & \textbf{PMV}    & \textbf{MOR}    & \textbf{LOS}    & \textbf{Avg}      \\
    \midrule
    \textit{BEEP} & \textit{59.43} & \textit{84.65} & \textit{72.71} & \textit{72.26} \\
    Our method  & 53.81 & 83.02 & 73.26 & 70.03 \\
    \bottomrule
    \end{tabular}
    \caption{AUROC scores in a subset of the clinical downstream document classification tasks. The macro-averaged AUROC score is calculated by taking the average of AUROC scores across this subset of tasks. The \textit{row in italic} indicates the model variant with the highest macro-averaged AUROC.}
    \label{tab:beep_results}
\end{table}

We compared our method with the state-of-the-art clinical outcome prediction model, BEEP~\cite{naikLiteratureAugmentedClinicalOutcome2022d}, which leverages a retrieval augmentation technique to enhance the predictive capabilities of clinical language models.
A small caveat is that BEEP focused on three downstream tasks: prolonged mechanical ventilation, mortality, and length of stay predictions.
We selected the best-performing solution from BEEP, UmlsBERT with weighted voting retrieval augmentation,  based on the averaged AUROC score to compare with our solution.
While BEEP outperforms our approach, particularly in the prediction of PMV, it is crucial to emphasise that our method achieves its predictions without relying on retrieval augmentation.
Future work may explore using retrieval augmentation on top of our proposed method.

\section{Training Configurations}
We use HuggingFace's Transformers~\cite{wolf-etal-2020-transformers} and PEFT~\cite{peft} libraries for the experiments.
All LLaMA-based models are trained on one NVIDIA A100-80GB GPU, while the baseline models are trained on a single NVIDIA GeForce GTX 1080 Ti-16GB GPU.

\section{Artefacts}

The pretrained baseline models including BioClinicalBERT~\cite{alsentzerPubliclyAvailableClinical2019}, BlueBERT~\cite{pengTransferLearningBiomedical2019}, and CORe~\cite{vanakenClinicalOutcomePrediction2021a} were released under the Creative Commons designation CC0 1.0 Universal license, whereas UmlsBERT~\cite{michalopoulosUmlsBERTClinicalDomain2021} was released under the MIT license.
LLaMA~\cite{touvronLLaMAOpenEfficient2023} was released under a noncommercial license.

MIMIC-III and MIMIC-IV dataset was released under the PhysioNet Credentialed Health Data License 1.5.0 and can only be accessed after one finishes the CITI Data or Specimens Only Research training\footnote{https://physionet.org/about/citi-course/}.

\end{document}